\documentclass[10pt,twocolumn,letterpaper]{article}

\usepackage{wacv}
\usepackage{times}
\usepackage{epsfig}
\usepackage{graphicx}
\usepackage{amsmath}
\usepackage{amssymb}
\usepackage{amsfonts}
\usepackage{algorithm}
\usepackage{algpseudocode}
\usepackage{booktabs}
\usepackage{graphics}
\usepackage{multirow}
\usepackage{tabularx}
\usepackage[accsupp]{axessibility}

\def\wacvPaperID{593} 

\wacvfinalcopy 

\ifwacvfinal
\def\assignedStartPage{9876} 
\fi

\ifwacvfinal
\usepackage[breaklinks=true,bookmarks=false]{hyperref}
\else
\usepackage[pagebackref=true,breaklinks=true,colorlinks,bookmarks=false]{hyperref}
\fi

\ifwacvfinal
\else
\fi

\begin{document}
\pagenumbering{gobble} 

\title{Self-Supervised Domain Adaptation for Visual Navigation \\with Global Map Consistency}

\author{
Eun Sun Lee \qquad Junho Kim \qquad Young Min Kim \\


Dept. of Electrical and Computer Engineering, Seoul National University, Korea\\
{\tt\small \{ eunsunlee, 82magnolia,  youngmin.kim \} @snu.ac.kr}
}

\maketitle

\begin{abstract}
We propose a light-weight, self-supervised adaptation for a visual navigation agent to generalize to unseen environment.
Given an embodied agent trained in a noiseless environment, our objective is to transfer the agent to a noisy environment where actuation and odometry sensor noise is present. 
Our method encourages the agent to maximize the consistency between the global maps generated at different time steps in a round-trip trajectory.
The proposed task is completely self-supervised, not requiring any supervision from ground-truth pose data or explicit noise model.
In addition, optimization of the task objective is extremely light-weight, as training terminates within a few minutes on a commodity GPU.
Our experiments show that the proposed task helps the agent to successfully transfer to new, noisy environments.
The transferred agent exhibits improved localization and mapping accuracy, further leading to enhanced performance in downstream visual navigation tasks. 
Moreover, we demonstrate test-time adaptation with our self-supervised task to show its potential applicability in real-world deployment.

\end{abstract}

\section{Introduction}

Vision-based navigation is a fundamental task for embodied AI agents that traverse the environment while conducting a series of actions.
The agent needs to plan and execute a sequence of movements based on its perception of the surrounding area to interact with the environment in a desirable way.
Training an agent in the real world is costly and thus, many research has made progress by leveraging simulators~\cite{habitat19iccv,szot2021habitat} utilizing 3D scene datasets~\cite{chang2017matterport3d,xiazamirhe2018gibsonenv}. 
While recent works demonstrate a nearly perfect performance with an agent trained in a noiseless simulator~\cite{wijmans2019dd}, the agent is bound to experience a critical decrease in its performance when deployed in a noisy, realistic environment~\cite{kadian2019we}. 
One may try to close the domain gap by fine-tuning the policy in a new, noisy environment~\cite{chaplot2020learning, kadian2019we}, but it is difficult to obtain ground truth labels or poses in a real-world deployment.
Furthermore, it is impossible to exactly model the various actuation and odometry sensor noise given the unlimited combinations of motors and sensors for custom robots deployed in various spatial layout and environmental configurations.

We investigate the scenario where the pre-trained model in a noiseless environment needs to be transferred to a new environment where the agent’s actuator is noisy and the odometry measurement is also erroneous.
As a mean to generalize in different environments, recent works propose modular approaches that separate the observation space and the action space ~\cite{chaplot2020object,gordon2018iqa, karkus2021differentiable}, as shown in Figure~\ref{fig:overview}. 
The observation module transfers the sensory perception into the domain-agnostic intermediate representation, and a subsequent policy module deduces the agent's action from the intermediate representation.
Then it suffices to fine-tune the observation module to produce the stable intermediate representation given corrupted measurements.
Still, adapting the observation module requires either ground-truth poses or explicit noise models, which are not available in our realistic set up.
%

To this end, we formulate a self-supervised learning task where the agent is encouraged to generate consistent global maps over a round-trip trajectory.
The agent generates the first global map during the forward trajectory.
After the turn point, the agent resets and generates a new map during the backward path from which we train the prediction of newly generated map.
The self-supervision signal compares the consistency between the overlapping maps generated from different trajectories and fine-tunes the pre-trained navigation agent to the noisy real world.
Our method does not require ground-truth poses as long as there exist overlapping observations of the global map.
After the observation module is fine-tuned to suppress the errors caused by the domain shift, we can combine it with the pre-trained action policy to perform various downstream tasks.

We extensively evaluate our self-supervised learning approach in fundamental navigation tasks such as localization, mapping, and exploration~\cite{chen2019learning} and demonstrate that our approach can quickly adapt the pre-trained agents to the new environment. Our proposed method finishes the adaptation process within 5 minutes, much faster than the domain randomization approach which converges after 65 minutes of fine-tuning. 
We further enhance the performance by data augmentation method utilizing random crops of trajectories.
Lastly, we train our agent in a test-time adaptation setting to demonstrate the task's applicability in the real world. 

In summary, our main contributions are as follows: i) we introduce a simple self-supervised task for robust visual navigation amidst actuation and sensor noise,
ii) we demonstrate that the suggested formulation enhances the localization, mapping, and the final task performance,  
iii) we show that our method is applicable for test-time adaptation indicating the potential applicability for real-world deployment.

\section{Related Work}

Our work proposes a self-supervised approach to transfer an embodied agent trained in a noiseless environment to a noisy environment where actuation and odometry sensor noise is present. 
We thus discuss relevant literature from the field of visual navigation for embodied AI, Sim2Real, and self-supervised learning.

\paragraph{Visual Navigation} 
In visual navigation for an embodied agent, the principal tasks are mapping and planning. 
Classical robotics approaches focus first on building a map of the environment~\cite{cadena2016past, durrant2006simultaneous, mur2015orb} followed by path planning~\cite{koenig2002d, wang2011application, frontier} on the map.
Recent deep-learning based approaches, on the other hand, often train an intelligent agent jointly for mapping and planning ~\cite{gupta2017cognitive, ramakrishnan2020occupancy}.
Such approaches have enabled to train the embodied agents for end-to-end tasks and adapt to a large number of challenging tasks including goal-oriented navigation~\cite{chaplot2020object, wani2020multion, wijmans2019dd}, audio-visual navigation~\cite{chen2019audio}, or embodied Q\&A~\cite{das2018embodied}. 

To perform the various navigational tasks, the intelligent agents utilize diverse forms of memory systems to represent the observed environment.
Neural networks can create implicit memory structure with LSTM or GRU~\cite{chen2019learning, mezghani2021memory} or a topological graph that aligns with location information such as landmarks or image frames on the graph nodes~\cite{ chaplot2020neural, chen2019behavioral,deng2020evolving, savinov2018semi, shah2020ving}.
However, the traditional map-based memories such as occupancy grid maps~\cite{chaplot2020learning,gupta2017cognitive, karkus2021differentiable, wani2020multion} still perform competitively, especially for long-range navigation tasks.
The explicit maps can also be easily extended into other map-based planning tasks and therefore widely deployed for visual navigation of embodied agents.
We utilize the occupancy map as the memory representation and the mean to define the self-supervision task.

\paragraph{Sim2Real}
When training an agent in a real-world setting, it is difficult to collect a large amount of data with ground-truth labels. 
Simulators are great tools for training embodied agents before deploying in the real world.
However, the policy learned from simulated environment often fails to maintain its performance in the real world setting~\cite{kadian2020sim2real}. 
Imitation learning observes the videos of desired trajectories collected by humans demonstration in the real environment~\cite{chen2019learning, das2018embodied}, but there still exists the gap between human and robot execution.
Thus, bridging the domain gap has become a new direction of research~\cite{chebotar2019closing, hansen2020self}.
Inspired by the domain randomization approach in visual domain~\cite{hansen2020self}, the approach from~\cite{Chattopadhyay2021RobustNavTB} randomizes the dynamic corruptions such as actuation and odometry noise. 
However, we demonstrate that our self-supervision approach can better compensate for the unknown noise in terms of both accuracy and efficiency, by transforming the observations into an intermediate representation~\cite{chen2020robust, li2020unsupervised}.

\paragraph{Self-Supervised Learning}
The major success of deep learning owes to a large amount of training data.
However, it is not scalable to collect the desired amount of labelled data for every task.
Self-supervised learning proposes to obtain labels from known information, namely auxiliary tasks, and apply techniques of supervised learning to extract useful representation for desired tasks.
Examples of such auxiliary tasks include predicting the random rotation of a given image or estimating depth and surface normals from observation frames ~\cite{Gidaris2018UnsupervisedRL, gordon2019splitnet, hansen2020self}.
With the careful design of auxiliary tasks, it has been demonstrated that the self-supervision can improve the performance of vision tasks such as 
image translation~\cite{park2020contrastive}, or video object segmentation~\cite{jabri2020walk}.
We suggest an orthogonal self-supervision task for various sensor noises, which could complement the aforementioned self-supervision for visual representation learning.
While previous studies on indoor navigation suggest selecting more reliable information from multi-modal sensor data~\cite{cai2020probabilistic, patel2019deep}, our proposed self-supervision does not utilize any additional sensors. 
Instead, we demonstrate that enforcing the consistency of the global maps is a useful auxiliary task to quickly adapt to the new sensor noise without ground truth labels.
\section{Method}

\begin{figure*}[!t]
\begin{center}
\includegraphics[width=2.0\columnwidth]{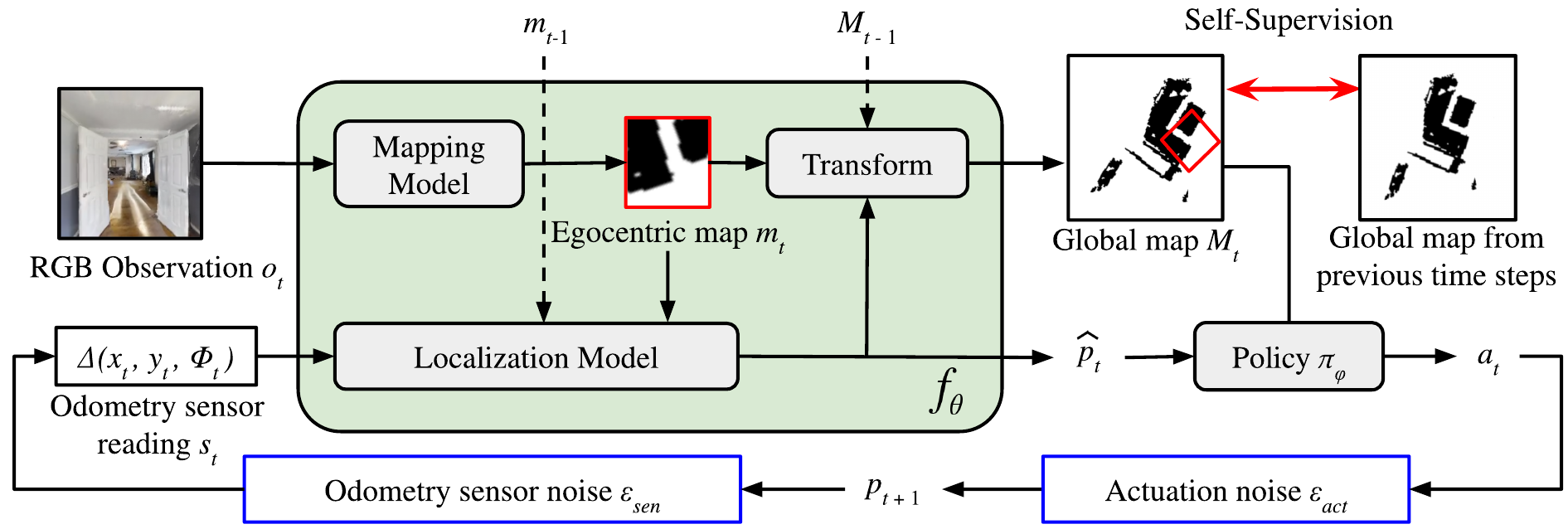}
\end{center}
\caption{\textbf{Overview of our approach.} 
Our domain adaption builds on a navigation agent that is composed of the mapping and localization module $f_\theta$ (shaded in green) and the planning module $\pi_\psi$.
Given an RGB observation $o_{t}$ and the odometry sensor reading $s_t$, $f_\theta$ estimates the pose $\hat{p}_t$ and the global map $M_t$, from which $\pi_\psi$ generates the task-specific policy to perform the action $a_t$.
After the mapping model predicts the egocentric map $m_t$ from the observation $o_t$, and the localization model estimates the agent's pose $\hat{p}_t$ by comparing the egocentric maps from two consecutive time\-steps, $m_{t-1}$ and $m_t$, in addition to reading the  odometry sensor $s_t$.
Then the generated map $m_t$ is transformed based on the estimated pose $\hat{p}_t$ and fused with the previous global map $M_{t-1}$ to update the global map $M_t$.
In a noiseless simulator, the agent can learn the parameters of the module $\theta$ from ground-truth $p_t$ and $m_t$.
In a real-world deployment with unknown actuation and odometry noise (outlined in blue), the ground-truth supervision is not obtainable. 
Our self-supervised learning approach transfers the agent to the noisy environment by computing the proposed global map consistency loss for the global map $M_t$. 
}
\label{fig:overview}
\end{figure*}

We introduce a simple and generic, self-supervised learning method for noise-robust visual navigation. 
Our objective is to transfer a pre\-trained agent to the novel, unseen environments where different types of actuation and odometry sensor noises are present. 
The proposed method enforces consistency on global maps generated over time in a round-trip trajectory. 
In this section, we explain the overall setup, the various types of actuation and sensor noise, and how the global map consistency is implemented. We also introduce our data augmentation method.

\subsection{Visual Navigation with Spatial Memory}
\label{sec:outline}

Our self-supervision works on the consistency of the global map, therefore it can be applied to most visual navigation agents that are based on spatial map memory~\cite{chaplot2020learning,chaplot2020object,karkus2021differentiable,ramakrishnan2020occupancy, wani2020multion,zhang2017neural}. 
Such approaches train two neural networks for mapping and planning, which we will refer as $f_\theta$ and $\pi_\psi$ respectively.

Figure~\ref{fig:overview} shows the overall flow of our approach. Given the current observation $o_t$ and the odometry sensor measurement $s_t$, the mapping model predicts the egocentric map $m_t$ and the localization model estimates the current position $\hat{p}_t$. 
Two models are jointly referred as the mapping and localization module $f_\theta$, where the global map $M_t$ is generated by combining $m_t$ using  $\hat{p}_t$.
Note that the predictions from the previous time step, the estimated egocentric map $m_{t-1}$ and global map $M_{t-1}$, are also utilized.   
Simply, we represent our mapping and localization module as  
\begin{equation}
f_{\theta}(o_{1:t},s_{1:t}) = M_{t}, \hat{p}_{t}.
\end{equation}
For the purpose of proposition, we demonstrate the applicability using a sequence of RGB image observations for $o_t$ as the input, and the 2D occupancy grid map for $M_t$ and the 2D pose for $p_t=(x,y,\phi)$.
Here $(x,y)$ denote the 2D coordinate and $\phi$ indicates the 1D orientation.

Then the planning module $\pi_\psi$ is implemented with a separate neural network, which is a policy network that observes the output of the mapping module and generates the action $a_t \in \mathcal{A}$
\begin{equation}
    \pi_{\psi}(M_{t},\hat{p}_{t}) = a_{t}.
\end{equation}
We allow a discrete set of agent actions, specifically $\mathcal{A}$=\{Forward, Turn Right, Turn Left\}.

Our supervision works on the intermediate representation between the two modules ($f_\theta$ and $\pi_\psi$), namely the consistency of $M_t$.
Note that the consistency of $M_t$ is tightly coupled with the accuracy of $\hat{p}_t$ because the pose $\hat{p}_t$ can be defined with respect to the generated map $M_t$ and the network $f_\theta$ fuses the current observation $o_t$ to $M_t$ using the current pose $\hat{p}_t$.
The formulation allows disentangling the variations in measurements from the policies designed for different downstream tasks. 
We can train a desired policy network $\pi_\psi$ in an ideal simulation setup with the ground truth oracle and fine-tune only the observation module $f_\theta$ in a different environment set-up. More specifically, in our setup, we only fine-tune the localization part of the module as the egocentric map prediction is not disturbed by the odometry or actuation noise.
Nevertheless, the proposed method is generic as it can be applied to various visual navigation models with distinct modules for perception and action sharing an intermediate representation.

\paragraph{Actuation and Odometry Sensor Noise} 
We first train both networks in a noiseless setup and demonstrate the feasibility of the proposed self-supervision in realistic noise models.
There are two main streams of expected variation; actuation and sensor noise.
A large amount of both classical and latest work~\cite{gonccalves2008sensor,khosla1989categorization, habitat19iccv} has presented the approach to model realistic actuation and odometry noise. 
In this section, we briefly describe the noise models used as an unseen environment for the pre-trained agent.
An exhaustive discussion is available in ~\cite{Chattopadhyay2021RobustNavTB}.

The actuation noise disturbs an agent's movement. 
The agent fails to reach the intended position, and it becomes difficult to estimate the agent's pose solely by the control commands.
The properties of actuation noise fall into two classes; motion bias, and motion drift. 
First, the motion bias adds a constant bias, $\delta_{c} \in \mathbb{R}^{3}$, and a stochastic bias, $\delta_{s} \in \mathbb{R}^{3}$ drawn from $ \mathcal{N}(\mu_s,\Sigma_{s})$.
Then, the motion drift, $\alpha \in \mathbb{R}^{3}$, is added towards left or right only for the forward movement. 
The actuation noise is the addition of both motion bias and motion drift
\begin{equation}
    \epsilon_{act} = \delta_{c} + \delta_{s} + \alpha. 
\end{equation}
The ground-truth pose of the agent after an action is defined as 
\begin{equation}
    (x_{t+1},y_{t+1},\phi_{t+1}) = (x_{t}, y_{t}, \phi_{t}) + (u_x,u_y,u_\phi) + \epsilon_{act},
    \label{eq:pose_actuation_noise}
\end{equation}
where the $(u_x,u_y,u_\phi)$ indicates the intended action control. 

In addition, the odometry sensor noise affects the pose estimation accuracy, which further degrades the mapping accuracy and the navigation performance.
The pose reading noise from odometry sensors is drawn from a Gaussian Mixture Model.
The final pose reading is measured as 
\begin{equation}
   (x_{t+1}',y_{t+1}',\phi_{t+1}') = (x_{t+1},y_{t+1},\phi_{t+1}) + k \epsilon_{sen},
   \label{eq:pose_odometry_noise}
\end{equation}
where the $k$ denotes the sensor noise severity.
The noise models are derived from the real-world observation, and we employ them based on the noise parameters modeled from actual physical deployment~\cite{locobot,murali2019pyrobot} to test the performance of our algorithm in a noisy environment.

\begin{figure}[t]
\begin{center}
\includegraphics[width=1.0\linewidth]{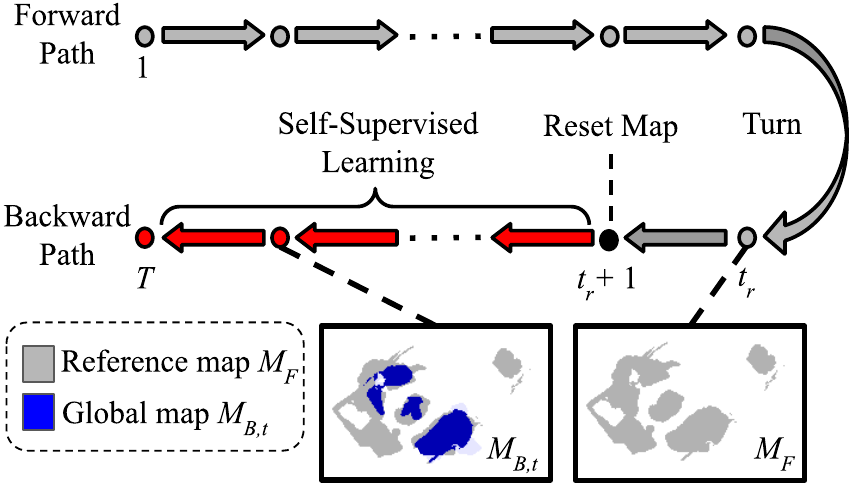}
\end{center}
   \caption{\textbf{Global Map Consistency from a Round Trip.} 
   The forward path from the time 1 to $t_r$ builds a reference map $M_F$. 
   At $t_r+1$, the agent resets the map and generates another map $M_{B,t}$ from $t_r+1$ to $T$. 
   At each step during the backward path, we impose the consistency between $M_F$ and $M_{B,t}$.}
\label{fig:task_overview}
\end{figure}

\subsection{Self Supervision with Global Map Consistency}
\label{sec:self-supervision}

We assume deploying a pre-trained agent in an unseen environment, where the ground truth poses or the changes in the observation noise model are not known.
Instead, we can create a supervision signal by enforcing consistency of the generated global map $M_t$.
While we cannot assure that the map is error-free, we can assume that the error accumulates over time.
This leads to incorporating more accurate pose data in generating the global map in the earlier steps compared to the ones generated later in a continuing trajectory. 

To implement the self-supervised learning task efficiently, we deliberately design overlapping trajectories and drive an embodied agent in round trips as shown in Figure~\ref{fig:task_overview}. 
The round trip is easily generated by executing any navigational task for a designated number of steps, turning around 180\textdegree, and executing the previous action sequence in reverse. 
The global map is first generated during the forward path, then the agent resets and generates another map from scratch during the backward path.
In a noiseless setting, the agent observes the same area at each one-way trip.  
Therefore, the global map from its forward path should be equal to the global map generated during its backward path.
When the actuation noise is present, the agent may not step on the same way\-points it has traversed during its forward path. Nonetheless, our proposed formulation is still valid as the agent generates maps from the overlapping area. 

To summarize, during the forward path, the agent generates a reference global map $M_{F}$
\begin{equation}
    f_{\theta}(o_{1:t_r},s_{1:t_r}) = M_{F}, \hat{p}_{F}, 
\end{equation}
where $t_{r}$ is the step when the robot turns around.
On its way back, the agent generates a new map.   
However, the same global coordinate is shared across all global maps and the agent continues incremental pose estimation using the full odometry $s_{1:t}$.
At each reverse step, a new global map, $M_{B,t}$, is predicted,
\begin{equation}
    f_{\theta}(o_{t_r+1:t},s_{1:t}) = M_{B,t}, \hat{p}_{B,t},\quad t>t_r.
\end{equation}

Lastly, our self-supervised task imposes the consistency between the predicted $M_{F}$ and $M_{B,t}$ to fine-tune the mapping module $f_\theta$ under the new environment. 
In the equation below, T denotes the ending time of the trajectory and $\mathtt{stopgrad}$ indicates that no back\-propagation should happen through $M_{F}$
\begin{equation}
    \mathcal{L} = \sum_{t=t_r+1}^{T}\|\mathtt{stopgrad}(M_{F}) - M_{B,t}\|_{2}.
    \label{eq:consistency}
\end{equation}
Therefore we compare the estimated map at every step during the backward path, providing a rich set for loss computations per acquired trajectory.
While there are several ways of finding the difference between the two occupancy grid maps, we incorporated the mean squared error (MSE) loss. 
The choice of losses is further discussed in the experiment section.

\begin{figure}[t]
\begin{center}
\includegraphics[width=1.0\linewidth]{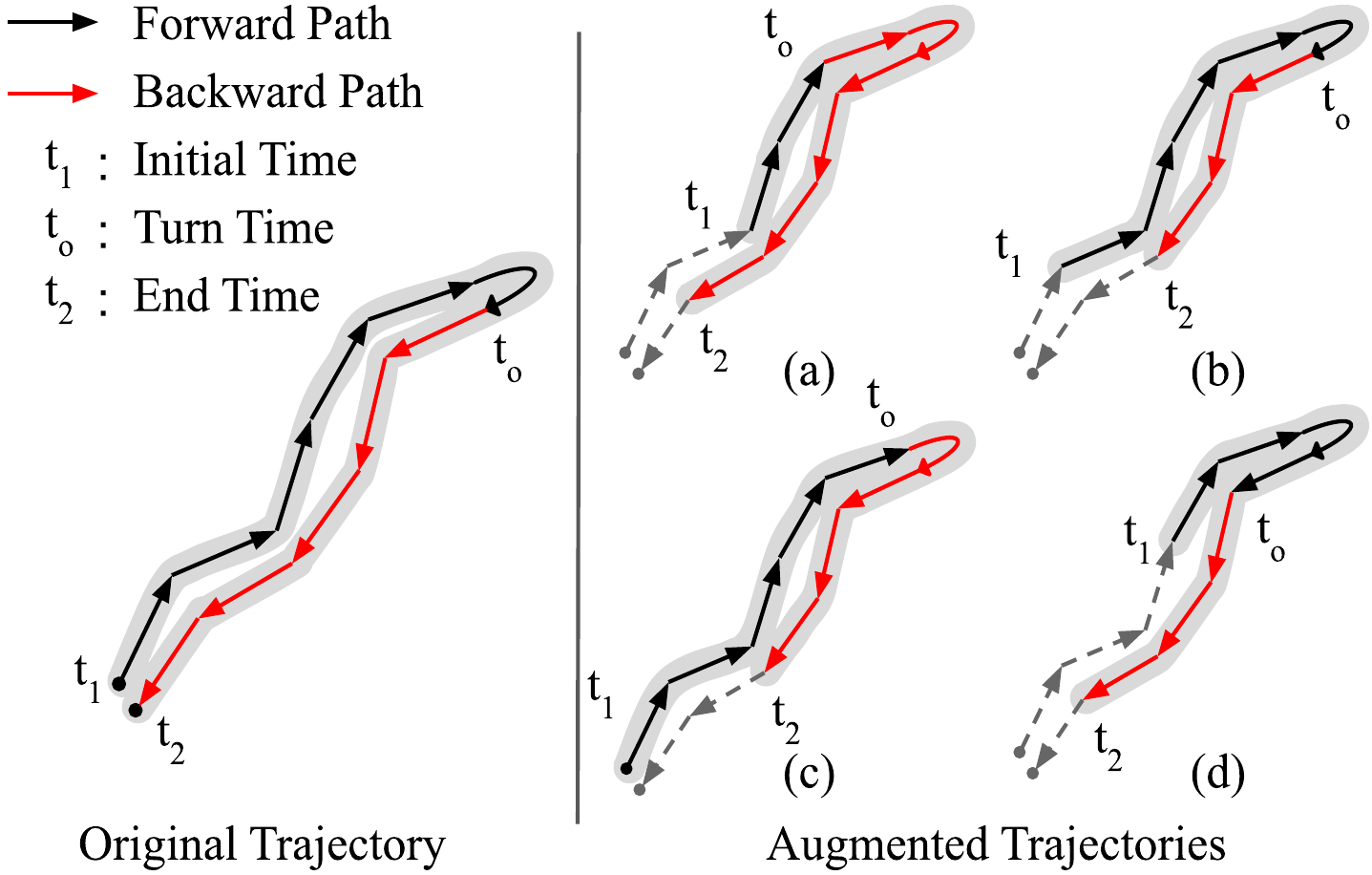}
\end{center}
   \caption{\textbf{Data Augmentation.} A round trip trajectory can be augmented by randomly sampling the three time steps, initial time, $t_1$, the turn time, $t_o$, and the end time, $t_2$, in a chronological order.}
\label{fig:data_augmentation}
\vspace{-1em}
\end{figure}

\paragraph{Data Augmentation}
While our vanilla formulation utilize the bisections of round-trip trajectories, the consistency can be enforced for any subsets of trajectories with overlapping observations.
We introduce a data augmentation method based on random cropping which has been used in various types of data such as videos or images~\cite{hansen2020self, takahashi2019data}.
Given a trajectory, we randomly sample three time\-steps in a chronological order: the initial time $t_1$, the turn time $t_o$, and the end time $t_2$ $(t_1 < t_o < t_2)$. A few example trajectories are shown in Figure~\ref{fig:data_augmentation}.
Then the reference map ${M}_F'$ is produced from  $f_{\theta}(o_{t_1:t_o},s_{t_1:t_o})$ and compared to the map ${M}_{B,t}'$ generated from the backward step $f_{\theta}(o_{t_o+1:t},s_{1:t})$, $t_o<t<t_2$.
Our loss is similarly defined as 
\begin{equation}
    \mathcal{L} =\sum_{t=t_{o}+1}^{t_{2}}\|\mathtt{stopgrad}({M}'_F)  - {M}'_{B,t}\|_{2}.
\end{equation}
We analyze the effectiveness of the proposed data augmentation method in the experiment section.

\section{Experiments}
 
We extensively evaluate the performance of a pre-trained embodied agent when adapted to a new environment with our self-supervised learning task.
In all our experiments we report the results using the agent from Active Neural SLAM~\cite{chaplot2020learning}, as it is a widely used baseline agent for visual navigation~\cite{chaplot2020object, chaplot2020neural,ramakrishnan2020occupancy}. 
Nonetheless, our proposed self-supervised learning task is generic as it is easily applicable to most navigation models or agents based on spatial representation. 
For navigation task analysis, we follow the task setup proposed by~\cite{chen2019learning}.
We evaluate the navigation agent by the area coverage within a fixed number of steps, which is 1000 steps in all of our experiments. 
We show our result for exploration since it is the fundamental task for most navigation agents~\cite{chaplot2020learning}.

We use Habitat~\cite{habitat19iccv} simulator in the scenes from Gibson dataset~\cite{xia2018gibson} and analyze on the standard train/val split~\cite{habitat19iccv} comparing against the ground truth poses and maps.
For our self-supervised learning approach, we adapt the original model~\cite{chaplot2020learning} implemented using PyTorch~\cite{paszke2019pytorch} to allow learning from global map consistency.
The fine-tuning time of the self-supervised learning, accelerated with an RTX 2080 GPU, completes within 5 minutes, and the test-time adaptation finishes in 2.5 minutes. 
Additional details about the experimental setup are available in the supplementary material.

\begin{figure*}[!t]
\begin{center}
\includegraphics[width=\textwidth]{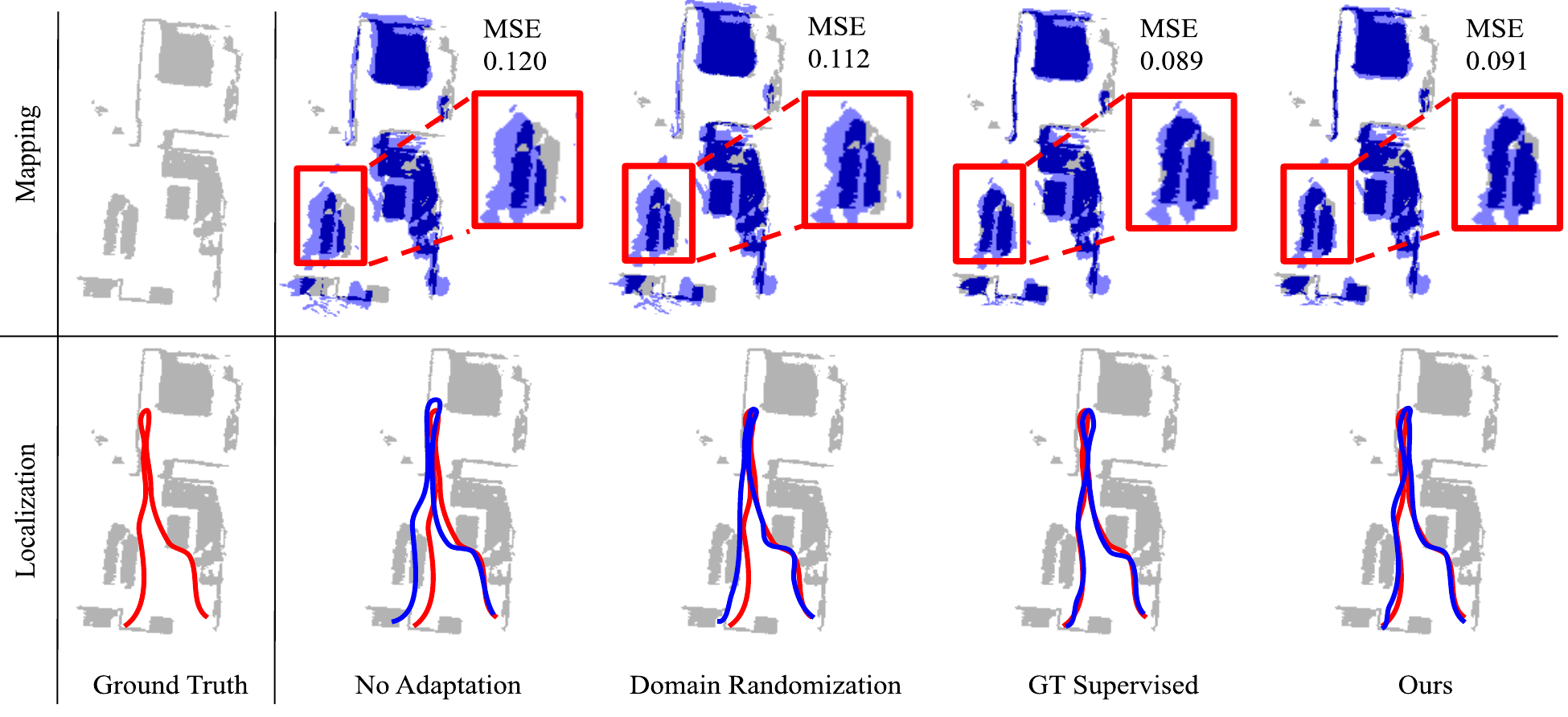}
\end{center}
   \caption{\textbf{Qualitative Result of Mapping and Localization.} The first row shows the qualitative result of the global maps generated from the same sequence of actions. The generated map is indicated in blue and the ground truth map is indicated in gray. 
   The second row shows the localization result where red is the ground-truth trajectory and blue is the estimated trajectory of each agent. 
   }
\label{fig:exp_map_traj}
\vspace{-1em}
\end{figure*}

\subsection{Task Adaptation to Noisy Environments}
\label{sec:main_result}

We first investigate if our proposed self-supervised learning task helps agents to transfer to a new, noisy environment. 
We pre-train an agent in a noiseless environment where the mapping and localization module $f_{\theta}$ is trained with the ground-truth pose and egocentric map.  
We then observe if our self-supervision can help the agent to generalize across various unseen noisy environments without ground-truth supervision. 
For unseen noisy environment, we apply the odometry noise and actuation noise models based on real data of LoCoBot~\cite{locobot} collected from the previous work ~\cite{chaplot2020learning,kadian2020sim2real}. 
For each episode, we first report the median localization error in terms of translation ($x,y$) and rotation ($\phi$), following ~\cite{campbell2018globally,campbell2019alignment}.
We additionally report the mean squared error (MSE) of the generated occupancy grid maps with respect to the ground-truth maps.
To quantify exploration performance, we report two metrics, namely the absolute coverage area in $\mbox{m}^2$ and the ratio of area coverage, similar to ~\cite{chaplot2020learning}.



\begin{table}[]
\begin{center}
\resizebox{\columnwidth}{!}{
\begin{tabular}{ll|cccc}
    \toprule
    \multicolumn{2}{l}{Method}& NA & GT & DR & Ours \\
    \midrule 
    \multirow{2}{*}{Localization} & $x,y$ (m) & 0.15 & 0.02 & 0.12 & 0.03 \\
     & $\phi$ ($^\circ$) & 2.67 & 0.27 & 1.7 & 0.36  \\
    \midrule
    Mapping & MSE & 0.25 & 0.17 & 0.23 & 0.17  \\
    \midrule
    \multirow{2}{*}{Exploration} & Ratio (\%) & 82.02 & 93.75 & 89.06 & 91.79 \\
    & Area ($\mbox{m}^2$) & 26.36 & 31.83 & 29.19 & 30.79 \\
    \midrule
    Training & Time (min) & - & - & 65 & 5 \\
    \bottomrule
\end{tabular} 
}
\end{center}
    \caption{Generalization to noisy environment evaluated on localization, mapping and exploration. Note that `MSE' denotes `mean squared error'.}

\label{table:generalization}
\vspace{-3ex}
\end{table} 

As seen in Table~\ref{table:generalization}, performance enhancement occurs in all evaluations.  
Our agent learns to better perceive the surroundings and explores 91.79\% of its environment when tested for exploration. 
``No Adaptation (\textbf{NA})" is the pre-trained model in a noiseless environment with ground-truth information, deployed in a new, noisy environment without any adaptation.
Thus, the errors in NA reflect the performance degradation due to the domain gap and sets the lower bound of adaption. 
Our agent distinctively outperforms the baseline agent and compensates for the unknown noise.
In contrast, ``GT Train (\textbf{GT})" is the model fine-tuned with the ground-truth supervision in the new environment.
While the ground-truth pose information is not available in the real world, GT serves as the upper bound of the performance for our adaptation model.
In addition to the enhancement in pose estimation, the self-supervised agent successfully generates occupancy representation as accurately as the GT does. 
We also compare our method to the ``Domain Randomization (\textbf{DR})"~\cite{Chattopadhyay2021RobustNavTB}, which exposes the model randomly to the various combinations of actuation and odometry noise and train with ground-truth supervision.    
To expose the agent with sufficient variations of unknown noise, DR needs a significant amount of training data ($\sim$1.5k trajectories) for fine-tuning.
Nonetheless, our model trained with the self-supervised learning method outperforms in all performance metrics by successfully compensating for the unknown noise within a faster training time.

The result exhibits that the global map consistency is a powerful self-supervision task for adapting a pre-trained agent to an unseen environment.
Figure~\ref{fig:exp_map_traj} shows the visualization of the global maps and pose trajectories from all agents executing the same sequence of actions.
The reconstructed map and the trajectory better align with the ground truth and therefore visually confirm that our model outperforms the pre-trained model and the domain randomization model.
More qualitative results are available in the supplementary material.
Below we also provide the analysis on various conditions of our proposed method.

\begin{figure*}[]
\centering
\includegraphics[width=\textwidth]{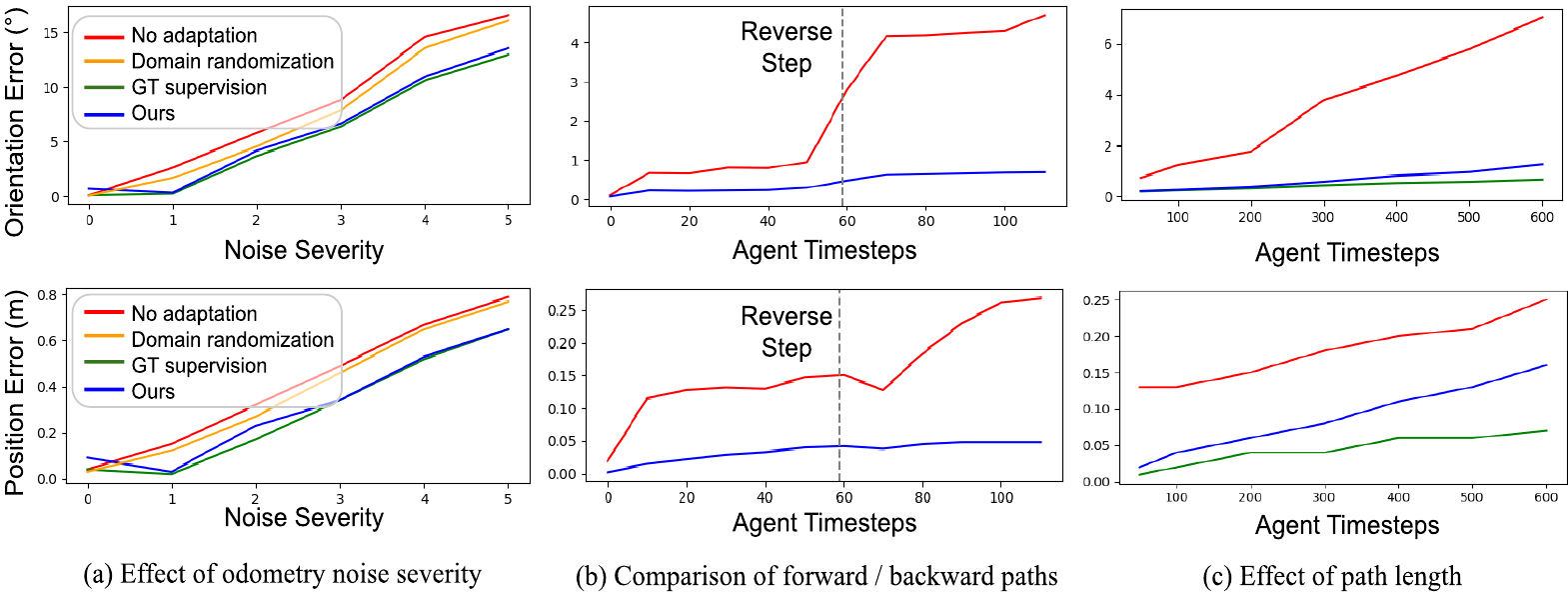}
   \caption{\textbf{Localization Error.} Performance analysis in task adaptation to noisy environments}
   
\label{fig:plot}
\vspace{-1em}
\end{figure*}


\begin{table}[]
\begin{center}
\resizebox{0.9\columnwidth}{!}{
\begin{tabular}{ll|cccc}
      \toprule 
      \multicolumn{2}{l}{Method} & NA & GT & DR & Ours \\
      \midrule 
      \multirow{2}{2cm}{Constant Motion Bias} & \text{$x,y$ (m)} & 0.16 & 0.03 & 0.14 & 0.04\\
         & \text{$\phi$ (\textdegree)} & 2.69 & 0.28  &1.93&  0.3\\
      \midrule
      \multirow{2}{2cm}{Stochastic Motion Bias} & \text{$x,y$ (m)} & 0.16 & 0.04 & 0.14 & 0.05\\
         & \text{$\phi$ (\textdegree)} & 2.77 & 0.32 & 2.06 &  0.34\\
      \midrule
      \multirow{2}{2cm}{Motion Drift} & \text{$x,y$ (m)} & 0.18 & 0.05  & 0.16 & 0.06\\
         & \text{$\phi$ (\textdegree)} & 3.91 & 0.49  & 2.39 & 0.82\\
      \bottomrule 
\end{tabular} 
}
\end{center}
  \caption{Localization error with respect to various types of actuation noise.}
  \label{Tab:actuationnoise}
  \vspace{-3ex}
\end{table}

\paragraph{Analysis on Various Noise Conditions}

We additionally demonstrate the performance of environment transfer under various noise conditions.
We first set up six environments where the sensor noise severity $k$ in Equation~\ref{eq:pose_odometry_noise} ranges from 0 to 5. 
Our agent is separately trained in each environment with the self-supervised learning task and compared against the original pre-trained model and the agent trained with ground-truth supervision.
Figure~\ref{fig:plot}(a) shows that our self-supervised agent consistently reports superior performance against the agent which is not fine-tuned. The result guarantees that the proposed self-supervised learning task can be utilized for various agents regardless of its odometry sensor noise level. 


Our self-supervision task does not assume any noise model and can be applied in various other types of noise.
To test the stability of the performance of our fine-tuned agent, we further set up three environments where the three types of actuation noise introduced in Section~\ref{sec:outline} are present, namely constant / stochastic motion bias and motion drift.
The outlined noise corruption reflects the actuation noise model from LoCoBot~\cite{kadian2019we,habitat19iccv} and it creates a more challenging environment for agents.
The result is reported in Table~\ref{Tab:actuationnoise}. 
In all cases, our transferred agent shows superior performance against the pre-trained model and exhibits high performance on a par with the GT Train model with the simple self-supervision regardless of unknown variations of noise models.

\vspace{-0.8em}
\paragraph{Forward vs. Backward Error}
Our formulation in Equation~\ref{eq:consistency} treats the map generated during the forward step $M_F$ as the supervision signal to backward map $M_{B,t}$.
However, $M_F$ is also error-prone, and one may question whether it is a valid assumption to treat the forward and backward path separately.
To answer the question, we analyze the accumulated pose errors in agent time steps during the round-trip trajectories. 
In Figure~\ref{fig:plot}(b), we observe that both position and orientation errors increase for all agents as they progress in a trajectory. 
The motion of turning steps induces significant movement compared to ordinary progression, and introduces a large amount of variation in both position and orientation, especially for the agent trained in a noiseless environment.
Although a similar trend is shown in our agent, the self-supervised learning task optimizes the agent to experience less variation.  
As shown in the graph, our agent consistently accumulates the pose estimation error within a similar extent during its forward and backward path. 
Therefore the backward path indeed suffers from comparably severe pose errors after the turning point, which could greatly be alleviated from the self-supervision.
\vspace{-0.8em}
\paragraph{Task Robustness to Path Length}
For our proposed self-supervision task, the only constraint we impose on the set of trajectories collected for training is round-trip.
We report that agents trained on a set of fixed-length round-trip trajectories can further generalize to longer trajectories.
In the standard setup presented in Table~\ref{table:generalization}, our proposed task fine-tunes the agent using short trajectories of 100 agent steps.
We show the adaptation performance when an agent is evaluated on trajectories of different lengths, varying from 50 steps up to 600. 
As the step length increases, we observe diminishing enhancement as shown in Figure~\ref{fig:plot}(c).
However, our proposed method still noticeably exceeds the lower bound performance across various trajectory lengths.
\vspace{-0.8em}
\paragraph{Task Robustness to Training Data Size}
We observe if the self-supervised learning task is trainable from a smaller set of collected trajectories. 
Our standard experiment uses 160 trajectories for training. 
In Table~\ref{Tab:datasize}, we show the result of our self-supervised agent trained on a different amount of trajectory data; 40, 80, 120, and 160.
It demonstrates the correlation that more number of trajectories leads to higher pose estimation and mapping performance. 
Nonetheless, an agent fine-tuned with 40 trajectories still achieves performance enhancement over the pre-trained agent.
Furthermore, we suggest an offline data augmentation method in Section~\ref{sec:self-supervision} which increases the number of trainable trajectories size under a limited set of conditions.

\begin{table}
\begin{center}
\resizebox{0.95\columnwidth}{!}{
\begin{tabular}{ll|ccccc} 
\toprule
      \multicolumn{2}{l}{Number of Trajectories} & NA & 40 & 80 & 120 & 160                      \\ 
\midrule
\multirow{2}{*}{Localization} & \text{$x,y$ (m)}   & 0.15 & 0.11  & 0.05  & 0.04   & 0.03                        \\
                      & $\phi$ (\textdegree)    & 2.67 & 1.47  & 0.4  & 0.38   & 0.36                        \\
\midrule
Mapping                   & \text{MSE}    & 0.25 & 0.23  & 0.18  & 0.18   & 0.17                        \\
\bottomrule
\end{tabular}
}
\end{center}
\caption{Robustness to training data size evaluated on localization and mapping. Note that `MSE' denotes `mean squared error'.}
\label{Tab:datasize}
\vspace{-2ex}
\end{table}

\subsection{Test-Time Adaptation}
\label{sec:test-time}

In the previous experiments, we followed the conventional train/val split in the Gibson dataset~\cite{xia2018gibson} and observed that the self-supervised learning task can adapt an agent to unknown actuation and odometry noise.
The training trajectories are collected from 72 training scenes and the performance is evaluated in 14 val scenes. 
The training scenes for fine-tuning were the same set of scenes used for the pre-training.
However, it is not realistic for an agent to observe the real-world noises in the pre-trained environment. 
In a practical scenario, a pre-trained agent might be expected to adapt to a new environment that is not available in a simulator.

For test-time adaptation, we design an experiment to demonstrate the task’s applicability for practical real-world deployment. 
We investigate the scenario where the pre-trained agent is fine-tuned in a single scene that has not been observed before.
The new scene contains noise identical to the standard experiment in Table~\ref{table:generalization}, which the agent was never exposed to during pre-training.
We also use a small set of trajectories and see if an agent can be fine-tuned without over-fitting.
We collect 20 trajectories for training and test the agent’s performance on 80 trajectories both within the same scene, compared to 160 trajectories used for training the agent presented in Section~\ref{sec:main_result}. 
The GT train agent is fine-tuned in a new scene with ground-truth data provided and tested in the same scene during 80 test trajectories. 

As shown in Table~\ref{Tab:testtime}, our agent performs highly accurate pose estimation and mapping. 
Moreover, in the exploration task result, our agent outperforms the GT train agent. 
Using only a small set of collected trajectories, our light-weight self-supervised learning finishes training an agent within 2.5 minutes.
Our result guarantees that the proposed task can quickly adapt agents in a real-world deployment setting.

\begin{table}[]
\begin{center}
\resizebox{0.83\columnwidth}{!}{
\begin{tabular}{ll|ccc}
      \toprule 
      \multicolumn{2}{l}{Method} & NA & GT & Ours\\
      \midrule 
      \multirow{2}{*}{Localization} & \text{$x,y$ (m)} & 0.15 & 0.02 & 0.05\\
        & $\phi$ (\textdegree) & 2.73 & 0.25 & 0.5\\
      \midrule
      \text{Mapping} & \text{MSE} & 0.16 & 0.11 & 0.12\\
      \midrule
      \multirow{2}{*}{Exploration} & \text{Ratio (\%)} & 82.02 & 91.06 & 92.25\\
        & Area ($m^2$) & 26.36 & 30.65 & 31.27\\
      \bottomrule 
\end{tabular}
}
\end{center}
  \caption{Performance analysis on test-time adaptation. }
 \vspace{-0.5em}

\label{Tab:testtime}
\end{table}

\begin{table}
\begin{center}
\resizebox{\columnwidth}{!}{
\begin{tabular}{l|ccc} 
\toprule
\multicolumn{1}{l}{Task} &\multicolumn{2}{c}{Localization} & Mapping     \\
\midrule
\multicolumn{1}{l}{Method} & $x,y$ (m) & $\phi$ (\textdegree)      & MSE  \\
\midrule
Ours     & 0.03   & 0.36            & 0.17       \\
Ours w/o data augmentation & 0.06   & 0.41            & 0.19       \\
Ours w/ last step update         & 0.15   & 2.57            & 0.24       \\
Ours w/ BCE Loss             & 0.16   & 2.78            & 0.26       \\
\bottomrule
\end{tabular}
}
\end{center}
\caption{Ablation Study}
\vspace{-1em}

\label{Tab:ablationstudy}
\end{table}

\subsection{Ablation study}
\label{sec:ablation}
In this section, we study ablated versions of our self-supervised learning method. 
All experiments are conducted in the same condition as described in Section~\ref{sec:main_result}.

\paragraph{Data Augmentation}
Given 160 trajectories collected with the pre-trained agent, we augment the data with the proposed random cropping method from Section~\ref{sec:self-supervision}. 
In Table ~\ref{Tab:ablationstudy}, we compare our agent against the agent trained without data augmentation. 
Our agent estimates the pose more accurately, showing that data augmentation helps agents to generalize across environments. 
\vspace{-0.8em}
\paragraph{Step-wise Supervision}
Our agent back-propagates the global map consistency loss at every step during the backward path. 
We examine if the step-wise supervision is necessary by comparing against the agent learns from the self-supervised learning only at its last step. 
The result presented in Table~\ref{Tab:ablationstudy} shows that training the self-supervised learning task at every step indeed is helpful and exhibits more accurate pose estimation and mapping compared to the ablated agent which trains only at the last step. 
\vspace{-0.8em}
\paragraph{MSE vs. BCE Loss} 
Another possible loss to compare the binary occupancy map is binary cross entropy (BCE) loss, which is used in many existing approaches for map prediction~\cite{chaplot2020learning,karkus2021differentiable}.
When used in our self-supervised task, BCE loss shows inferior performance in pose estimation than MSE.
We empirically observe that BCE loss incurs noisy gradients compared to MSE loss, with higher variance in magnitude.
Such instability in gradients leads to poor convergence and deteriorates fine-tuning.
Further analysis on gradients of the two losses is in the supplementary material.

\section{Conclusion}
In this paper, we presented a light-weight and simple, self-supervised learning task for agents transferring to new, noisy environments. 
We demonstrated the quick and robust adaptation performance of the self-supervised agent to various actuation and odometry sensor noise on a par with the agent trained with ground-truth data.  
Our result presents an insight that the global map consistency provides a better understanding of the agent's pose and its surroundings. 
We also observe that improvement in observation space leads to the enhancement in action space for embodied visual navigation tasks. 
We expect our proposed method to be easily applicable to most visual navigation agents due to its simplicity. 
Furthermore, our optimistic test-time adaptation result points at the real-world deployment as the future research direction.

\section*{Acknowledgements}
This work was supported by the National Research Foundation of Korea(NRF) grant funded by the Korea government(MSIT) (No. NRF-2020R1C1C1008195), Samsung Electronics Co., Ltd, and the BK21 FOUR program of the Education and Research Program for Future ICT Pioneers, Seoul National University in 2021.

{\small
\bibliographystyle{ieee_fullname}
\bibliography{egbib}

\begin{thebibliography}{10}\itemsep=-1pt

\bibitem{locobot}
{\em "LoCoBot: An open source low cost robot"}.
\newblock \url{http://www.locobot.org/}, 2021.

\bibitem{cadena2016past}
Cesar Cadena, Luca Carlone, Henry Carrillo, Yasir Latif, Davide Scaramuzza,
  Jos{\'e} Neira, Ian Reid, and John~J Leonard.
\newblock Past, present, and future of simultaneous localization and mapping:
  Toward the robust-perception age.
\newblock {\em IEEE Transactions on robotics}, 32(6):1309--1332, 2016.

\bibitem{cai2020probabilistic}
Peide Cai, Sukai Wang, Yuxiang Sun, and Ming Liu.
\newblock Probabilistic end-to-end vehicle navigation in complex dynamic
  environments with multimodal sensor fusion.
\newblock {\em IEEE Robotics and Automation Letters}, 5(3):4218--4224, 2020.

\bibitem{campbell2018globally}
Dylan Campbell, Lars Petersson, Laurent Kneip, and Hongdong Li.
\newblock Globally-optimal inlier set maximisation for camera pose and
  correspondence estimation.
\newblock {\em IEEE transactions on pattern analysis and machine intelligence},
  42(2):328--342, 2018.

\bibitem{campbell2019alignment}
Dylan Campbell, Lars Petersson, Laurent Kneip, Hongdong Li, and Stephen Gould.
\newblock The alignment of the spheres: Globally-optimal spherical mixture
  alignment for camera pose estimation.
\newblock In {\em Proceedings of the IEEE/CVF Conference on Computer Vision and
  Pattern Recognition}, pages 11796--11806, 2019.

\bibitem{chang2017matterport3d}
Angel Chang, Angela Dai, Thomas Funkhouser, Maciej Halber, Matthias Niessner,
  Manolis Savva, Shuran Song, Andy Zeng, and Yinda Zhang.
\newblock Matterport3d: Learning from rgb-d data in indoor environments.
\newblock {\em arXiv preprint arXiv:1709.06158}, 2017.

\bibitem{chaplot2020learning}
Devendra~Singh Chaplot, Dhiraj Gandhi, Saurabh Gupta, Abhinav Gupta, and Ruslan
  Salakhutdinov.
\newblock Learning to explore using active neural slam.
\newblock {\em arXiv preprint arXiv:2004.05155}, 2020.

\bibitem{chaplot2020object}
Devendra~Singh Chaplot, Dhiraj~Prakashchand Gandhi, Abhinav Gupta, and Russ~R
  Salakhutdinov.
\newblock Object goal navigation using goal-oriented semantic exploration.
\newblock {\em Advances in Neural Information Processing Systems}, 33, 2020.

\bibitem{chaplot2020neural}
Devendra~Singh Chaplot, Ruslan Salakhutdinov, Abhinav Gupta, and Saurabh Gupta.
\newblock Neural topological slam for visual navigation.
\newblock In {\em Proceedings of the IEEE/CVF Conference on Computer Vision and
  Pattern Recognition}, pages 12875--12884, 2020.

\bibitem{Chattopadhyay2021RobustNavTB}
Prithvijit Chattopadhyay, Judy Hoffman, R. Mottaghi, and Aniruddha Kembhavi.
\newblock Robustnav: Towards benchmarking robustness in embodied navigation.
\newblock {\em ArXiv}, abs/2106.04531, 2021.

\bibitem{chebotar2019closing}
Yevgen Chebotar, Ankur Handa, Viktor Makoviychuk, Miles Macklin, Jan Issac,
  Nathan Ratliff, and Dieter Fox.
\newblock Closing the sim-to-real loop: Adapting simulation randomization with
  real world experience.
\newblock In {\em 2019 International Conference on Robotics and Automation
  (ICRA)}, pages 8973--8979. IEEE, 2019.

\bibitem{chen2020robust}
Bryan Chen, Alexander Sax, Gene Lewis, Iro Armeni, Silvio Savarese, Amir Zamir,
  Jitendra Malik, and Lerrel Pinto.
\newblock Robust policies via mid-level visual representations: An experimental
  study in manipulation and navigation.
\newblock {\em arXiv preprint arXiv:2011.06698}, 2020.

\bibitem{chen2019audio}
Changan Chen, Unnat Jain, Carl Schissler, Sebastia Vicenc~Amengual Gari, Ziad
  Al-Halah, Vamsi~Krishna Ithapu, Philip Robinson, and Kristen Grauman.
\newblock Audio-visual embodied navigation.
\newblock {\em environment}, 97:103, 2019.

\bibitem{chen2019behavioral}
Kevin Chen, Juan~Pablo de Vicente, Gabriel Sepulveda, Fei Xia, Alvaro Soto,
  Marynel V{\'a}zquez, and Silvio Savarese.
\newblock A behavioral approach to visual navigation with graph localization
  networks.
\newblock {\em arXiv preprint arXiv:1903.00445}, 2019.

\bibitem{chen2019learning}
Tao Chen, Saurabh Gupta, and Abhinav Gupta.
\newblock Learning exploration policies for navigation.
\newblock {\em arXiv preprint arXiv:1903.01959}, 2019.

\bibitem{das2018embodied}
Abhishek Das, Samyak Datta, Georgia Gkioxari, Stefan Lee, Devi Parikh, and
  Dhruv Batra.
\newblock Embodied question answering.
\newblock In {\em Proceedings of the IEEE Conference on Computer Vision and
  Pattern Recognition}, pages 1--10, 2018.

\bibitem{deng2020evolving}
Zhiwei Deng, Karthik Narasimhan, and Olga Russakovsky.
\newblock Evolving graphical planner: Contextual global planning for
  vision-and-language navigation.
\newblock {\em arXiv preprint arXiv:2007.05655}, 2020.

\bibitem{durrant2006simultaneous}
Hugh Durrant-Whyte and Tim Bailey.
\newblock Simultaneous localization and mapping: part i.
\newblock {\em IEEE robotics \& automation magazine}, 13(2):99--110, 2006.

\bibitem{Gidaris2018UnsupervisedRL}
Spyros Gidaris, Praveer Singh, and N. Komodakis.
\newblock Unsupervised representation learning by predicting image rotations.
\newblock {\em ArXiv}, abs/1803.07728, 2018.

\bibitem{gonccalves2008sensor}
Jos{\'e} Gon{\c{c}}alves, Jos{\'e} Lima, H{\'e}lder Oliveira, and Paulo Costa.
\newblock Sensor and actuator modeling of a realistic wheeled mobile robot
  simulator.
\newblock In {\em 2008 IEEE International Conference on Emerging Technologies
  and Factory Automation}, pages 980--985. IEEE, 2008.

\bibitem{gordon2019splitnet}
Daniel Gordon, Abhishek Kadian, Devi Parikh, Judy Hoffman, and Dhruv Batra.
\newblock Splitnet: Sim2sim and task2task transfer for embodied visual
  navigation.
\newblock In {\em Proceedings of the IEEE/CVF International Conference on
  Computer Vision}, pages 1022--1031, 2019.

\bibitem{gordon2018iqa}
Daniel Gordon, Aniruddha Kembhavi, Mohammad Rastegari, Joseph Redmon, Dieter
  Fox, and Ali Farhadi.
\newblock Iqa: Visual question answering in interactive environments.
\newblock In {\em Proceedings of the IEEE conference on computer vision and
  pattern recognition}, pages 4089--4098, 2018.

\bibitem{gupta2017cognitive}
Saurabh Gupta, James Davidson, Sergey Levine, Rahul Sukthankar, and Jitendra
  Malik.
\newblock Cognitive mapping and planning for visual navigation.
\newblock In {\em Proceedings of the IEEE Conference on Computer Vision and
  Pattern Recognition}, pages 2616--2625, 2017.

\bibitem{hansen2020self}
Nicklas Hansen, Rishabh Jangir, Yu Sun, Guillem Aleny{\`a}, Pieter Abbeel,
  Alexei~A Efros, Lerrel Pinto, and Xiaolong Wang.
\newblock Self-supervised policy adaptation during deployment.
\newblock {\em arXiv preprint arXiv:2007.04309}, 2020.

\bibitem{jabri2020walk}
Allan Jabri, Andrew Owens, and Alexei~A Efros.
\newblock Space-time correspondence as a contrastive random walk.
\newblock {\em Advances in Neural Information Processing Systems}, 2020.

\bibitem{kadian2019we}
Abhishek Kadian, Joanne Truong, Aaron Gokaslan, Alexander Clegg, Erik Wijmans,
  Stefan Lee, Manolis Savva, Sonia Chernova, and Dhruv Batra.
\newblock Are we making real progress in simulated environments? measuring the
  sim2real gap in embodied visual navigation.
\newblock 2019.

\bibitem{kadian2020sim2real}
Abhishek Kadian, Joanne Truong, Aaron Gokaslan, Alexander Clegg, Erik Wijmans,
  Stefan Lee, Manolis Savva, Sonia Chernova, and Dhruv Batra.
\newblock Sim2real predictivity: Does evaluation in simulation predict
  real-world performance?
\newblock {\em IEEE Robotics and Automation Letters}, 5(4):6670--6677, 2020.

\bibitem{karkus2021differentiable}
Peter Karkus, Shaojun Cai, and David Hsu.
\newblock Differentiable slam-net: Learning particle slam for visual
  navigation.
\newblock In {\em Proceedings of the IEEE/CVF Conference on Computer Vision and
  Pattern Recognition}, pages 2815--2825, 2021.

\bibitem{khosla1989categorization}
Pradeep~K Khosla.
\newblock Categorization of parameters in the dynamic robot model.
\newblock {\em IEEE Transactions on Robotics and Automation}, 5(3):261--268,
  1989.

\bibitem{koenig2002d}
Sven Koenig and Maxim Likhachev.
\newblock D\^{}* lite.
\newblock {\em Aaai/iaai}, 15, 2002.

\bibitem{li2020unsupervised}
Shangda Li, Devendra~Singh Chaplot, Yao-Hung~Hubert Tsai, Yue Wu,
  Louis-Philippe Morency, and Ruslan Salakhutdinov.
\newblock Unsupervised domain adaptation for visual navigation.
\newblock {\em arXiv preprint arXiv:2010.14543}, 2020.

\bibitem{mezghani2021memory}
Lina Mezghani, Sainbayar Sukhbaatar, Thibaut Lavril, Oleksandr Maksymets, Dhruv
  Batra, Piotr Bojanowski, and Karteek Alahari.
\newblock Memory-augmented reinforcement learning for image-goal navigation.
\newblock {\em arXiv preprint arXiv:2101.05181}, 2021.

\bibitem{mur2015orb}
Raul Mur-Artal, Jose Maria~Martinez Montiel, and Juan~D Tardos.
\newblock Orb-slam: a versatile and accurate monocular slam system.
\newblock {\em IEEE transactions on robotics}, 31(5):1147--1163, 2015.

\bibitem{murali2019pyrobot}
Adithyavairavan Murali, Tao Chen, Kalyan~Vasudev Alwala, Dhiraj Gandhi, Lerrel
  Pinto, Saurabh Gupta, and Abhinav Gupta.
\newblock Pyrobot: An open-source robotics framework for research and
  benchmarking.
\newblock {\em arXiv preprint arXiv:1906.08236}, 2019.

\bibitem{park2020contrastive}
Taesung Park, Alexei~A Efros, Richard Zhang, and Jun-Yan Zhu.
\newblock Contrastive learning for unpaired image-to-image translation.
\newblock In {\em European Conference on Computer Vision}, pages 319--345.
  Springer, 2020.

\bibitem{paszke2019pytorch}
Adam Paszke, Sam Gross, Francisco Massa, Adam Lerer, James Bradbury, Gregory
  Chanan, Trevor Killeen, Zeming Lin, Natalia Gimelshein, Luca Antiga, et~al.
\newblock Pytorch: An imperative style, high-performance deep learning library.
\newblock {\em Advances in neural information processing systems},
  32:8026--8037, 2019.

\bibitem{patel2019deep}
Naman Patel, Anna Choromanska, Prashanth Krishnamurthy, and Farshad Khorrami.
\newblock A deep learning gated architecture for ugv navigation robust to
  sensor failures.
\newblock {\em Robotics and Autonomous Systems}, 116:80--97, 2019.

\bibitem{ramakrishnan2020occupancy}
Santhosh~K Ramakrishnan, Ziad Al-Halah, and Kristen Grauman.
\newblock Occupancy anticipation for efficient exploration and navigation.
\newblock In {\em European Conference on Computer Vision}, pages 400--418.
  Springer, 2020.

\bibitem{savinov2018semi}
Nikolay Savinov, Alexey Dosovitskiy, and Vladlen Koltun.
\newblock Semi-parametric topological memory for navigation.
\newblock {\em arXiv preprint arXiv:1803.00653}, 2018.

\bibitem{habitat19iccv}
Manolis Savva, Abhishek Kadian, Oleksandr Maksymets, Yili Zhao, Erik Wijmans,
  Bhavana Jain, Julian Straub, Jia Liu, Vladlen Koltun, Jitendra Malik, Devi
  Parikh, and Dhruv Batra.
\newblock Habitat: {A} {P}latform for {E}mbodied {AI} {R}esearch.
\newblock In {\em Proceedings of the IEEE/CVF International Conference on
  Computer Vision (ICCV)}, 2019.

\bibitem{shah2020ving}
Dhruv Shah, Benjamin Eysenbach, Gregory Kahn, Nicholas Rhinehart, and Sergey
  Levine.
\newblock Ving: Learning open-world navigation with visual goals.
\newblock {\em arXiv preprint arXiv:2012.09812}, 2020.

\bibitem{szot2021habitat}
Andrew Szot, Alex Clegg, Eric Undersander, Erik Wijmans, Yili Zhao, John
  Turner, Noah Maestre, Mustafa Mukadam, Devendra Chaplot, Oleksandr Maksymets,
  Aaron Gokaslan, Vladimir Vondrus, Sameer Dharur, Franziska Meier, Wojciech
  Galuba, Angel Chang, Zsolt Kira, Vladlen Koltun, Jitendra Malik, Manolis
  Savva, and Dhruv Batra.
\newblock Habitat 2.0: Training home assistants to rearrange their habitat.
\newblock {\em arXiv preprint arXiv:2106.14405}, 2021.

\bibitem{takahashi2019data}
Ryo Takahashi, Takashi Matsubara, and Kuniaki Uehara.
\newblock Data augmentation using random image cropping and patching for deep
  cnns.
\newblock {\em IEEE Transactions on Circuits and Systems for Video Technology},
  30(9):2917--2931, 2019.

\bibitem{wang2011application}
Huijuan Wang, Yuan Yu, and Quanbo Yuan.
\newblock Application of dijkstra algorithm in robot path-planning.
\newblock In {\em 2011 second international conference on mechanic automation
  and control engineering}, pages 1067--1069. IEEE, 2011.

\bibitem{wani2020multion}
Saim Wani, Shivansh Patel, Unnat Jain, Angel~X Chang, and Manolis Savva.
\newblock Multion: Benchmarking semantic map memory using multi-object
  navigation.
\newblock {\em arXiv preprint arXiv:2012.03912}, 2020.

\bibitem{wijmans2019dd}
Erik Wijmans, Abhishek Kadian, Ari Morcos, Stefan Lee, Irfan Essa, Devi Parikh,
  Manolis Savva, and Dhruv Batra.
\newblock Dd-ppo: Learning near-perfect pointgoal navigators from 2.5 billion
  frames.
\newblock {\em arXiv preprint arXiv:1911.00357}, 2019.

\bibitem{xiazamirhe2018gibsonenv}
Fei Xia, Amir R.~Zamir, Zhi-Yang He, Alexander Sax, Jitendra Malik, and Silvio
  Savarese.
\newblock Gibson {Env}: real-world perception for embodied agents.
\newblock In {\em Computer Vision and Pattern Recognition (CVPR), 2018 IEEE
  Conference on}. IEEE, 2018.

\bibitem{xia2018gibson}
Fei Xia, Amir~R Zamir, Zhiyang He, Alexander Sax, Jitendra Malik, and Silvio
  Savarese.
\newblock Gibson env: Real-world perception for embodied agents.
\newblock In {\em Proceedings of the IEEE Conference on Computer Vision and
  Pattern Recognition}, pages 9068--9079, 2018.

\bibitem{frontier}
B. Yamauchi.
\newblock A frontier-based approach for autonomous exploration.
\newblock In {\em Proceedings 1997 IEEE International Symposium on
  Computational Intelligence in Robotics and Automation CIRA'97. 'Towards New
  Computational Principles for Robotics and Automation'}, pages 146--151, 1997.

\bibitem{zhang2017neural}
Jingwei Zhang, Lei Tai, Ming Liu, Joschka Boedecker, and Wolfram Burgard.
\newblock Neural slam: Learning to explore with external memory.
\newblock {\em arXiv preprint arXiv:1706.09520}, 2017.

\end{thebibliography}
}

\end{document}